\documentclass[pdflatex]{nolta}
\usepackage{mathtools}
\usepackage{bm}
\usepackage[ruled,vlined]{algorithm2e}

\Vol{17}%
\No{1}%
\Year{2026}%
\Month{1}%
\PaperID{2026ENP0886}

\title{Efficient event-driven retrieval in high-capacity kernel Hopfield networks}

\AUTHOR{ \author{Akira Tamamori}{1}\orcid{0009-0000-8893-0058} }

\AFFILIATE{ \affiliate{Faculty of Information Science, Aichi
    Institute of Technology\\1247 Yachigusa, Yakusa-cho, Toyota-shi, Aichi 470-0392, Japan}{1} }

\received{10}{27}{20XX}
\revised{12}{29}{20XX}
\published{7}{1}{20XX}

\begin{document}

\begin{abstract}
  High-capacity associative memory models, such as Kernel Logistic
  Regression (KLR) Hopfield networks, have demonstrated strong storage
  capabilities but typically rely on computationally expensive
  synchronous updates. This reliance poses a bottleneck for deployment
  on energy-efficient, event-driven neuromorphic hardware. In this
  paper, we investigate the asynchronous retrieval dynamics of KLR
  Hopfield networks. We show empirically that, under appropriately
  tuned kernel parameters, asynchronous sequential updates exhibit
  trajectories that are statistically indistinguishable from those of
  synchronous dynamics, while maintaining high recall accuracy within
  the tested regime for random patterns. Furthermore, we find that the
  asynchronous network achieves empirical storage capacities
  approaching $P/N \approx 30$ in static random pattern regimes,
  exceeding classical limits. To evaluate computational efficiency, we
  analyze the total number of state transitions (bit flips) required
  for error correction. The results show that the network converges
  using a number of events close to the initial Hamming distance from
  the target pattern, without observable spurious oscillations. These
  findings suggest that the large-margin attractors induced by KLR
  learning create a smooth energy landscape suited for sparse,
  event-driven computation, providing a basis for scalable and
  low-power associative memory on neuromorphic architectures.
\end{abstract}

\maketitle

\section{Introduction}
\label{sec:intro}
Associative memory models, typified by the Hopfield network, are
fundamental architectures for pattern retrieval and error correction
in neural computing. Recent studies have shown that integrating Kernel
Logistic Regression (KLR) into the learning process significantly
enhances the storage capacity of these networks.  While classical
Hopfield models are bounded by severe storage limits, including the
statistical mechanics bound of $P\approx 0.14N$~\cite{Amit1985} and
strict information-theoretic constraints on error-free
retrieval~\cite{McEliece1987}, KLR networks have been shown to
reliably store patterns at loads exceeding $P/N > 4.0$ in static
memory tasks~\cite{tamamori_letter, tamamori_nolta_a,
  tamamori_nolta_b}.

To overcome classical capacity limits, a prominent contemporary
approach involves Modern Hopfield Networks (MHNs) or Dense Associative
Memories~\cite{Krotov2016, Ramsauer2021}. These architectures
significantly extend the state-of-the-art by modifying the energy
function to include steep non-linearities (e.g., exponential
interactions), achieving substantially increased, and theoretically
exponential, storage capacities~\cite{Demircigil2017}.  However, their
reliance on complex, non-local activation functions, such as Softmax,
presents significant challenges for direct implementation on highly
parallel, energy-efficient neuromorphic hardware. In contrast, the KLR
Hopfield networks investigated in this study retain a simpler,
binary-state architecture with a standard quadratic energy structure
in the feature space. By shifting the burden of capacity enhancement
from architectural redesign to the optimization process (margin
maximization), our approach provides a scalable and hardware-friendly
alternative.

Despite these theoretical and empirical successes, the practical
deployment of high-capacity kernel associative memories on
resource-constrained hardware faces a fundamental bottleneck. The
conventional retrieval dynamics of KLR networks rely on synchronous
updates, which require evaluating the input potential for all $N$
neurons simultaneously at each time step. In the context of kernel
methods, this involves computing a dense matrix-vector multiplication
over all $P$ stored patterns. For large-scale applications where $P$
is substantially larger than $N$, this synchronous update scheme
incurs substantial computational and memory access costs.

To address this computational bottleneck, neuromorphic engineering
often employs asynchronous, event-driven architectures~\cite{Roy2019,
  Davies2018, Furber2014}. In these systems, such as Spiking Neural
Networks (SNNs)~\cite{Maass1997}, computation and communication are
triggered only when the state of a specific neuron
changes~\cite{Pfeiffer2018, Merolla2014}, an approach highly effective
for processing sparse, asynchronous data streams~\cite{Gallego2020}.
However, the applicability of asynchronous updates to high-capacity
kernel networks remains largely unexplored. A fundamental issue is the
classical trade-off between storage capacity and the size of the
basins of attraction~\cite{Amit1985, McEliece1987}. As the number of
stored patterns $P$ increases, the attractor basins inevitably tend to
shrink, rendering the system more susceptible to initial noise and
variations in the update trajectory.  Therefore, a critical concern is
whether asynchronous, sequential updates would introduce spurious
oscillations or severely degrade the noise robustness and capacity of
the KLR network under such highly constrained, high-load conditions.

In this paper, we systematically investigate the asynchronous
retrieval dynamics of KLR Hopfield networks. We present empirical
evidence that, under optimized kernel parameters, the asynchronous
update scheme provides an efficient and functionally comparable
alternative to synchronous dynamics. Our main contributions are
threefold:
\begin{enumerate}
\item We empirically establish the equivalence of synchronous and
  asynchronous retrieval dynamics in KLR networks. Under the tested
  conditions, the convergence trajectories are nearly identical,
  suggesting that the underlying energy landscape is smooth and robust
  against perturbations in the update order, even when the basins are
  narrowed due to high storage loads.

\item We empirically compare the retrieval trajectories of synchronous
  and asynchronous dynamics in KLR networks. Under the tested
  conditions, the convergence trajectories are nearly
  indistinguishable within statistical variation, suggesting that the
  underlying energy landscape is smooth and robust against
  perturbations in the update order.
  
\item We evaluate the empirical storage capacity limit of the network
  in the static memory regime. By optimizing the kernel locality
  parameter for robust retrieval, we demonstrate that the network can
  maintain perfect recall accuracy at storage loads approaching
  $P/N \approx 30$, far exceeding previous estimates.

\item We quantify the computational efficiency of the asynchronous
  update scheme. Our results indicate an efficient retrieval
  trajectory; the total number of required state updates (bit flips)
  closely matches the initial Hamming distance from the target
  pattern. This minimal-event dynamic indicates that KLR networks are
  well-suited for event-driven neuromorphic hardware implementations,
  offering substantial reductions in computational cost while
  preserving capacity.
\end{enumerate}

The remainder of this paper is organized as
follows. Section~\ref{sec:methods} details the network model and the
update schemes. Section~\ref{sec:similarity} presents the comparative
analysis of retrieval trajectories. Section~\ref{sec:capacity}
evaluates the storage capacity limit. Section~\ref{sec:efficiency}
evaluates the efficiency of the event-driven updates. Finally,
Section~\ref{sec:discussion} discusses the implications for hardware
design, and Section~\ref{sec:conclusion} concludes the paper.

\section{Model and Methods}
\label{sec:methods}
In this section, we define the Kernel Logistic Regression (KLR)
Hopfield network model~\cite{tamamori_nolta_a, tamamori_nolta_b}. We
then describe the two update schemes evaluated in this study, namely
synchronous and asynchronous dynamics, and define the metrics used to
quantify retrieval performance and computational efficiency.

\subsection{Kernel Logistic Regression Hopfield Network}
We consider an associative memory network of $N$ neurons, whose state
is represented by a bipolar vector $\bm{s} \in \{-1, 1\}^N$, that
stores $P$ patterns $\{\boldsymbol{\xi}^\mu\}_{\mu=1}^P$. The
retrieval dynamics are governed by the local field $h_i(\bm{s})$ at
neuron $i$:
\begin{equation}
    h_i(\bm{s}) = \sum_{\mu=1}^P \alpha_{\mu i} K(\bm{s}, \boldsymbol{\xi}^\mu),
    \label{eq:local_field}
\end{equation}
where $\boldsymbol{\alpha} \in \mathbb{R}^{P \times N}$ is the matrix
of dual variables learned via KLR~\cite{Schölkopf2001, Hastie2009},
and $K(\cdot, \cdot)$ is a kernel function. Throughout this study, we
use the RBF kernel,
$K(\bm{x}, \bm{y}) = \exp(-\gamma \|\bm{x} - \bm{y}\|^2)$, with the
locality parameter $\gamma$.

For completeness, we briefly describe the KLR learning objective. The
dual variables for each neuron $i$, denoted as
$\boldsymbol{\alpha}_i \in \mathbb{R}^P$, are learned independently by
minimizing an $L_2$-regularized negative log-likelihood. The objective
function $L(\boldsymbol{\alpha}_i)$ is defined as:
\begin{align}
  L(\boldsymbol{\alpha}_i) &= - \sum_{\nu=1}^{P} \left[ y_{\nu,i} \log(\sigma(h_i(\boldsymbol{\xi}^\nu))) \right. \nonumber\\
                           &+ \left.(1 - y_{\nu,i}) \log(1 - \sigma(h_i(\boldsymbol{\xi}^\nu))) \right] + \frac{\lambda}{2}\boldsymbol{\alpha}_i^{\top} \bm{K}\boldsymbol{\alpha}_i, 
  \label{eq:loss_function}
\end{align}
where $y_{\nu,i} = (\xi_i^\nu + 1)/2 \in \{0, 1\}$ is the target bit,
$\sigma(z) = 1/(1+e^{-z})$ denotes the logistic sigmoid function,
$\bm{K}$ is the Gram matrix whose elements are given by
$K_{\mu\nu} = K(\boldsymbol{\xi}^\mu, \boldsymbol{\xi}^\nu)$, and
$\lambda$ is the weight decay parameter. This optimization yields
large-margin attractors that are central to our analysis.

\subsection{Update Schemes and Energy Landscape}
The evolution of the network state $\bm{s}(t)$ depends on the update
schedule. To analyze the stability of these dynamics, we define a
pseudo-energy function (or a Lyapunov function candidate)
$V(\bm{s})$~\cite{tamamori_nolta_b} that characterizes the alignment
between the network state and the local field:
\begin{equation}
  V(\bm{s}) = -\sum_{i=1}^N s_i h_i(\bm{s}) = -\sum_{i=1}^N s_i \sum_{\mu=1}^P \alpha_{\mu i} K(\bm{s}, \boldsymbol{\xi}^\mu).
\label{eq:pseudo_energy}
\end{equation}
Note that we use $V(\bm{s})$ to denote our pseudo-energy,
distinguishing it from the Ising energy $E(\bm{s})$ of classical
Hopfield networks, which guarantees a monotonic decrease under
asynchronous updates.  We consider two distinct update schemes and
their implications for this energy landscape.

\textbf{1. Synchronous (Parallel) Update:} In the synchronous scheme,
all $N$ neurons update their states simultaneously based on the
current state $\bm{s}(t)$:
\begin{equation}
s_i(t+1) = \text{sign}(h_i(\bm{s}(t))) \quad \text{for all } i = 1, \dots, N.
\label{eq:sync_update}
\end{equation}
(We adopt the convention that $\text{sign}(0)=1$.) While this parallel
update is computationally efficient on GPUs, it does not guarantee a
monotonic decrease of the pseudo-energy $V(\bm{s})$. Consequently,
synchronous dynamics in recurrent neural networks are theoretically
susceptible to oscillatory behavior. Depending on the network
structure and initial conditions, synchronous updates can converge to
stable periodic orbits of various periods, or even exhibit complex
trajectories~\cite{Saito2024}. This susceptibility is particularly
relevant under high storage loads, where the energy landscape becomes
highly frustrated.

\textbf{2. Asynchronous (Sequential) Update:} In the asynchronous
scheme, neurons are updated sequentially. At each update step, a
single neuron $i$ is selected, and its state is updated based on the
current network state:
\begin{equation}
s_i^{\text{new}} = \text{sign}(h_i(\bm{s}^{\text{current}})).
\label{eq:async_update}
\end{equation}
Unlike the synchronous case, this sequential update rule suppresses
macroscopic oscillations. While $V(\bm{s})$ is not a strict Lyapunov
function for general asymmetric kernels, the large classification
margins induced by KLR learning drive the local field $h_{i}$ to align
with the stable state, ensuring that individual bit flips tend to
decrease the pseudo-energy in practice (see
Appendix~\ref{app:energy_landscape} for a detailed mechanistic
discussion). Therefore, the asynchronous dynamics tend to converge to
a fixed-point attractor.

For a fair comparison with the synchronous scheme, we define one
\textit{epoch} of asynchronous dynamics as $N$ sequential updates,
where the update order is determined by a random permutation of the
indices $\{1,\ldots,N\}$.
\subsection{Performance and Efficiency Metrics}
To evaluate the network, we consider both its capacity and the
computational cost of retrieval.

\begin{itemize}
\item \textbf{Pattern Recall Accuracy:} The proportion of noisy
  initial states that converge to their respective target patterns
  ($\bm{s}^{\text{final}} = \boldsymbol{\xi}^{\mu}$). This metric
  quantifies the storage capacity and noise robustness.
\item \textbf{Event Count (Bit Flips):} To quantify the computational
  cost in an event-driven setting, we track the total number of
  \textit{events} during the retrieval process. An event is defined as
  a state transition where the updated state differs from the current
  state ($s_{i}^{\text{new}} \neq s_{i}^{\text{current}}$). In an
  ideal event-driven hardware implementation, computations are
  triggered only when such a bit flip occurs.
\end{itemize}

\subsection{Experimental Setup}
Unlike our previous studies that focused on the highly localized
``Ridge'' regime ($\gamma=0.02$) to maximize local stability for
static memory~\cite{tamamori_nolta_a}, the present study explores a
broader ``Robust'' regime ($\gamma=0.1$). As shown in the following
sections, this slightly larger kernel width provides the wider
attractor basins necessary for stable retrieval under high noise
levels and large storage loads, while still maintaining high
capacity. The dual variables $\boldsymbol{\alpha}$ are learned using
standard gradient descent with a learning rate of $\eta=0.1$, a weight
decay of 0.01, and a fixed number of 500 update iterations.

To ensure the statistical reliability, all performance metrics
(accuracy and event counts) are averaged over 50 independent trials,
each with different random pattern realizations and noise
configurations. For the retrieval dynamics, an initial noise level is
introduced by randomly flipping a specified percentage (e.g., 10\% or
20\%) of the bits in a target pattern.

All simulations were implemented in Python 3.13 using the \verb+NumPy+
2.1.3 and \verb+SciPy+ 1.15.2 libraries and were executed on a
standard workstation equipped with an Intel Core i9-9900K processor
and 64 GB of RAM. No GPU acceleration was used, as the primary focus
of this study is the algorithmic efficiency of the retrieval dynamics
rather than large-scale parallel training.

\subsection{Algorithmic Implementation}
\begin{algorithm}[t]
\caption{KLR Training via Gradient Descent}
\label{alg:training}
\SetAlgoLined
\DontPrintSemicolon
\textbf{Input:} Training patterns $\boldsymbol{\Xi} = \{\boldsymbol{\xi}^1, \dots, \boldsymbol{\xi}^P\}$, kernel parameter $\gamma$, learning rate $\eta$, weight decay $\lambda$, max iterations $T_{\text{max}}$.\\
\textbf{Output:} Learned dual variables $\boldsymbol{\alpha} \in \mathbb{R}^{P \times N}$.\\
\textbf{Initialize:} $\boldsymbol{\alpha} \leftarrow \bm{0}_{P \times N}$\\
Compute Gram matrix: $K_{\mu\nu} = \exp(-\gamma \|\boldsymbol{\xi}^\mu - \boldsymbol{\xi}^\nu\|^2)$ for $\mu, \nu \in \{1,\dots,P\}$\\
Transform targets to binary $\{0, 1\}$: $Y_{\mu i} = (\xi_i^\mu + 1) / 2$\\
\For{$t = 1$ \KwTo $T_{\text{max}}$}{
    Compute logits: $\bm{Z} \leftarrow \bm{K} \boldsymbol{\alpha}$ \tcp*{$\bm{Z} \in \mathbb{R}^{P \times N}$}
    Compute probabilities: $\bm{P} \leftarrow 1 / (1 + \exp(-\bm{Z}))$ \tcp*{Sigmoid}
    Compute gradient: $\nabla L \leftarrow \bm{K}^\top (\bm{P} - \bm{Y}) + \lambda \boldsymbol{\alpha}$\\
    Update weights: $\boldsymbol{\alpha} \leftarrow \boldsymbol{\alpha} - \eta \nabla L$\\
}
\Return $\boldsymbol{\alpha}$
\end{algorithm}

\begin{algorithm}[t]
\caption{Asynchronous Event-Driven Retrieval}
\label{alg:retrieval}
\SetAlgoLined
\DontPrintSemicolon
\textbf{Input:} Initial state $\bm{s} \in \{-1, 1\}^N$, learned $\boldsymbol{\alpha}$, stored patterns $\boldsymbol{\Xi}$, kernel $\gamma$, max epochs $E_{\text{max}}$.\\
\textbf{Output:} Retrieved state $\bm{s}$, total event count $C_{\text{event}}$.\\
\textbf{Initialize:} $C_{\text{event}} \leftarrow 0$\\
\For{$e = 1$ \KwTo $E_{\text{max}}$}{
    $\bm{s}_{\text{old}} \leftarrow \bm{s}$\\
    Generate a random permutation $\pi$ of indices $\{1, \dots, N\}$\\
    \For{each $i \in \pi$}{
        Compute local field for neuron $i$ based on \textit{current} state $\bm{s}$:\\
        $h_i \leftarrow \sum_{\mu=1}^P \alpha_{\mu i} \exp(-\gamma \|\bm{s} - \boldsymbol{\xi}^\mu\|^2)$\\
        Determine new state: $s_i^{\text{new}} \leftarrow \text{sign}(h_i)$ \tcp*{Assume $\text{sign}(0)=1$}
        \If{$s_i^{\text{new}} \neq s_i$}{
            $\bm{s}_i \leftarrow s_i^{\text{new}}$ \tcp*{Update state immediately}
            $C_{\text{event}} \leftarrow C_{\text{event}} + 1$ \tcp{Record an event}
        }
    }
    \If{$\bm{s} = \bm{s}_{\text{old}}$}{
        \textbf{break} \tcp*{Converged to a fixed point}
    }
}
\Return $\bm{s}$, $C_{\text{event}}$
\end{algorithm}

To ensure reproducibility and clarify the computational procedures, we
provide the pseudocodes for the KLR training phase
(Algorithm~\ref{alg:training}) and the asynchronous retrieval phase
(Algorithm~\ref{alg:retrieval}).

\section{Empirical Similarity of Synchronous and Asynchronous Dynamics}
\label{sec:similarity}
In classical Hopfield networks, synchronous (parallel) updates are
prone to oscillatory behavior or convergence to spurious limit cycles,
particularly under high storage loads. Asynchronous (sequential)
updates guarantee convergence to a fixed point by ensuring a monotonic
decrease in energy, but often follow different, potentially
sub-optimal trajectories compared to the synchronous case. To evaluate
the suitability of KLR networks for event-driven hardware, we first
examine whether asynchronous updates degrade retrieval performance or
alter the underlying dynamics.

\begin{figure}[t]
\centering
\includegraphics[width=\columnwidth]{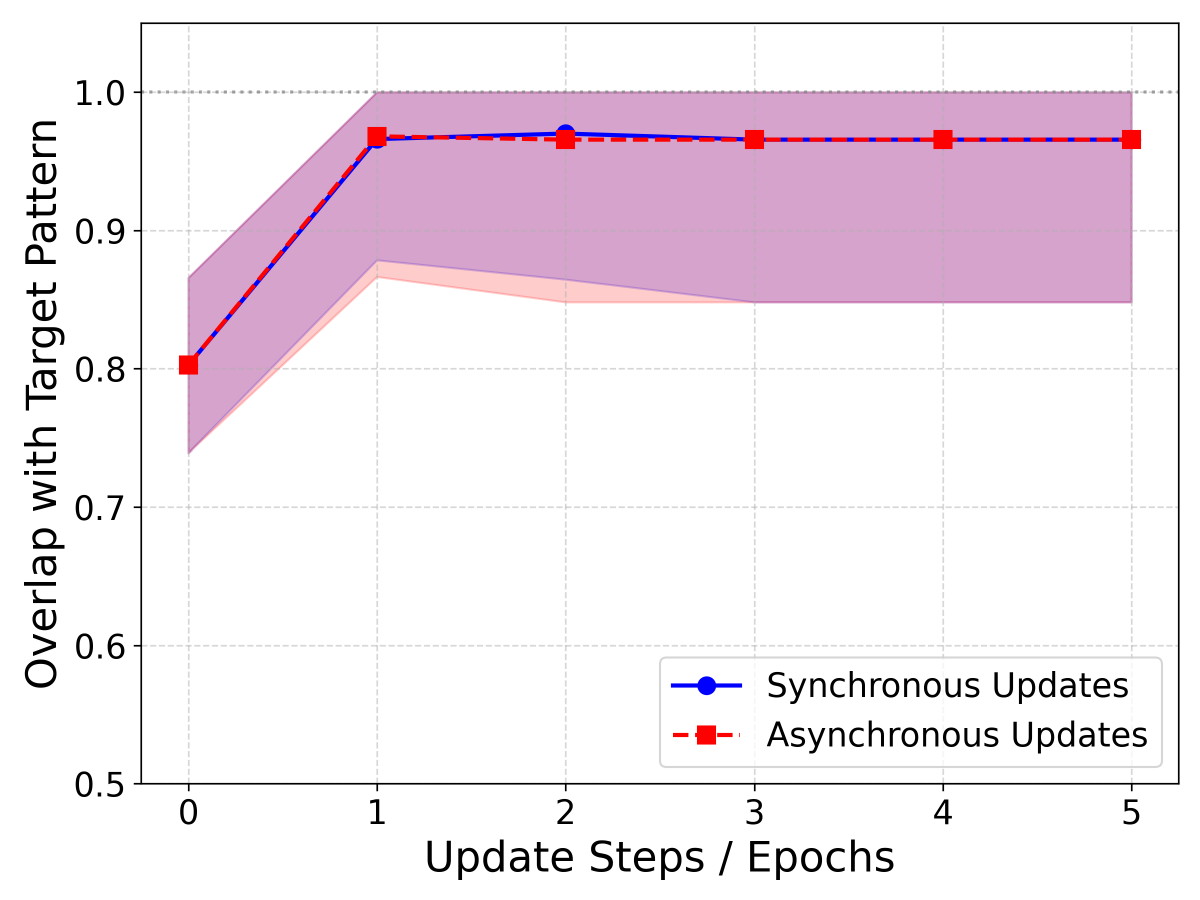}
\caption{Retrieval dynamics under synchronous and asynchronous
  updates. The plot compares the convergence trajectory of synchronous
  (blue solid line) and asynchronous (red dashed line) updates from an
  initial state with 20\% noise ($N=50$, $P/N=3.0$, $\gamma=0.1$).
  Shaded areas indicate standard deviation over 50 trials.  (Note: The
  upper bound of the shaded regions is clipped at 1.0, as the overlap
  cannot physically exceed this value.)  Under these conditions, both
  schemes converge to a high recall state along nearly
  indistinguishable trajectories within statistical variation.}
\label{fig:retrieval_dynamics}
\end{figure}

We simulated the retrieval process starting from a noisy initial
state, where 20\% of the bits in a target pattern were randomly
flipped. Figure~\ref{fig:retrieval_dynamics} plots the average overlap
between the network state and the target pattern as a function of the
update steps (for synchronous) or epochs (for asynchronous).

The results indicate a clear empirical similarity between the two
update schemes. Both the synchronous and asynchronous dynamics exhibit
a smooth, monotonic increase in overlap, converging to high recall
(Overlap $>$ 0.95) within a few steps. Importantly, under our tested
conditions, no significant difference is observed between the two
trajectories within statistical variation, and the variance across 50
independent trials (shaded areas) remains small.  To confirm that this
empirical equivalence is not an artifact of small network sizes, we
repeated these experiments for larger networks ($N=100$ and
$N=200$). As detailed in Appendix~\ref{app:scalability_dynamics}, the
synchronous and asynchronous trajectories remain nearly
indistinguishable at these larger scales.

This observation is non-trivial. It suggests that, for random patterns
and optimal kernel parameters, the energy landscape induced by KLR
learning appears to be smooth and to lack pronounced rugged local
minima that could trap a sequentially updated state. Regardless of the
random permutation order used in the asynchronous epochs, the network
appears to be consistently driven toward the correct attractor
basin. This provides empirical evidence that implementing KLR networks
on asynchronous, event-driven hardware can preserve the retrieval
performance observed in software simulations, without introducing
significant order-dependent artifacts.

\section{Empirical Storage Capacity Limit}
\label{sec:capacity}
Having established the empirical similarity of retrieval trajectories
under moderate loads, we next investigate the scalability and the
empirical storage capacity limit of the network for static random
patterns. Classical Hopfield networks with Hebbian learning are
theoretically bounded by a capacity of
$P \approx 0.14N$~\cite{Amit1985}. Exceeding this limit results in a
rapid proliferation of spurious states and a failure of pattern
retrieval. To evaluate the capability and scalability of KLR networks,
we systematically increased the storage load $P/N$ while keeping the
initial noise level at 10\%. To address potential finite-size effects,
we conducted this analysis across three network sizes: $N=50$, $100$,
and $200$.

\begin{figure}[t]
    \centering
    \includegraphics[width=\columnwidth]{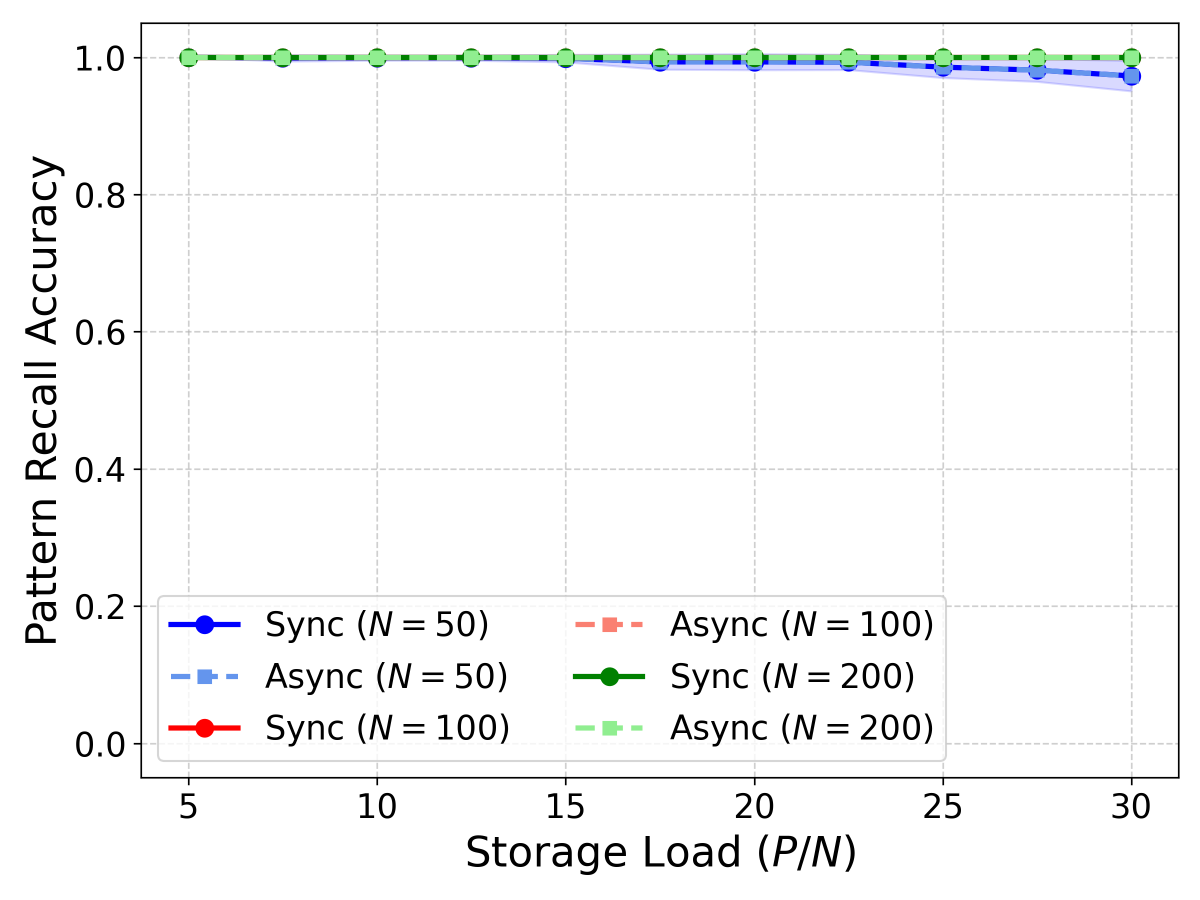}
    \caption{Capacity scaling under synchronous and asynchronous
      updates. The pattern recall accuracy is plotted against the
      storage load ($P/N$) under 10\% initial noise
      ($\gamma=0.1$). Results for three network sizes are shown:
      $N=50$ (blue), $N=100$ (red), and $N=200$ (green). Solid and
      dashed lines represent synchronous and asynchronous updates,
      respectively, with shaded areas indicating the standard
      deviation over 50 trials. Note that the solid and dashed lines
      largely overlap for all sizes. Furthermore, the red lines
      ($N=100$) are fully overlaid by the green lines ($N=200$), as
      both maintain an accuracy of 1.0 up to $P/N=30$.}
    \label{fig:capacity_scaling}
\end{figure}

Figure~\ref{fig:capacity_scaling} presents the pattern recall accuracy
of both synchronous and asynchronous update schemes as a function of
the storage load $P/N$. The networks were trained using the robust
kernel parameter $\gamma=0.1$. The results, averaged over 50
independent trials, indicate a high resilience to memory congestion
that scales positively with network size.

The results demonstrate that the network maintains a recall accuracy
of 1.0 up to high loads. For $N=50$, the performance begins to degrade
around $P/N \approx 20$. However, for larger networks ($N=100, 200$),
the accuracy remains at 1.0 even at $P/N = 30$ (the maximum load
tested). This corresponds to a storage capacity that is orders of
magnitude higher than the classical theoretical limit within the
tested finite-size regime, providing a lower bound for the capacity
under these conditions, which benefits from the increasing
orthogonality of random patterns in the high-dimensional kernel
feature space.

A key finding is that the recall accuracy of the synchronous (solid
lines) and asynchronous (dashed lines) schemes are consistent across
all network sizes and storage loads. The error bands overlap within
statistical variance, supporting the functional similarity of the two
dynamics even when the energy landscape becomes highly constrained by
strong interference from thousands of stored patterns.

This scaling analysis provides empirical evidence that the high
capacity of KLR networks is not a small-scale artifact, but a property
that persists across the tested system sizes. Furthermore, it confirms
that the deep attractor basins formed by the kernel map are
sufficiently robust to guide individual, sequential neuron updates to
the correct attractor, supporting the feasibility of deploying these
high-capacity networks on asynchronous, event-driven hardware
platforms at scale.

\section{Efficiency of Event-Driven Retrieval}
\label{sec:efficiency}
In the previous sections, we established that asynchronous updates
preserve the empirical storage capacity and noise robustness of KLR
networks. However, the primary motivation for adopting asynchronous
dynamics is the reduction of computational cost. In an event-driven
hardware architecture (e.g., SNNs), computation is sparse in time; a
neuron only consumes energy and updates its state when it receives a
spike (event) from a presynaptic neuron whose state has changed.

To quantify the computational efficiency of the asynchronous KLR
network, we measured the total number of state transitions, or
\textit{events} (bit flips), required to converge to the target
pattern from a noisy initial state. We compare this actual event count
with a theoretical lower bound: the initial Hamming distance between
the noisy state and the target pattern. Any bit flip beyond this
minimum indicates an inefficient trajectory, potentially involving
oscillations or convergence to intermediate spurious states before
finding the true attractor.

\begin{figure}[t]
\centering
\includegraphics[width=\columnwidth]{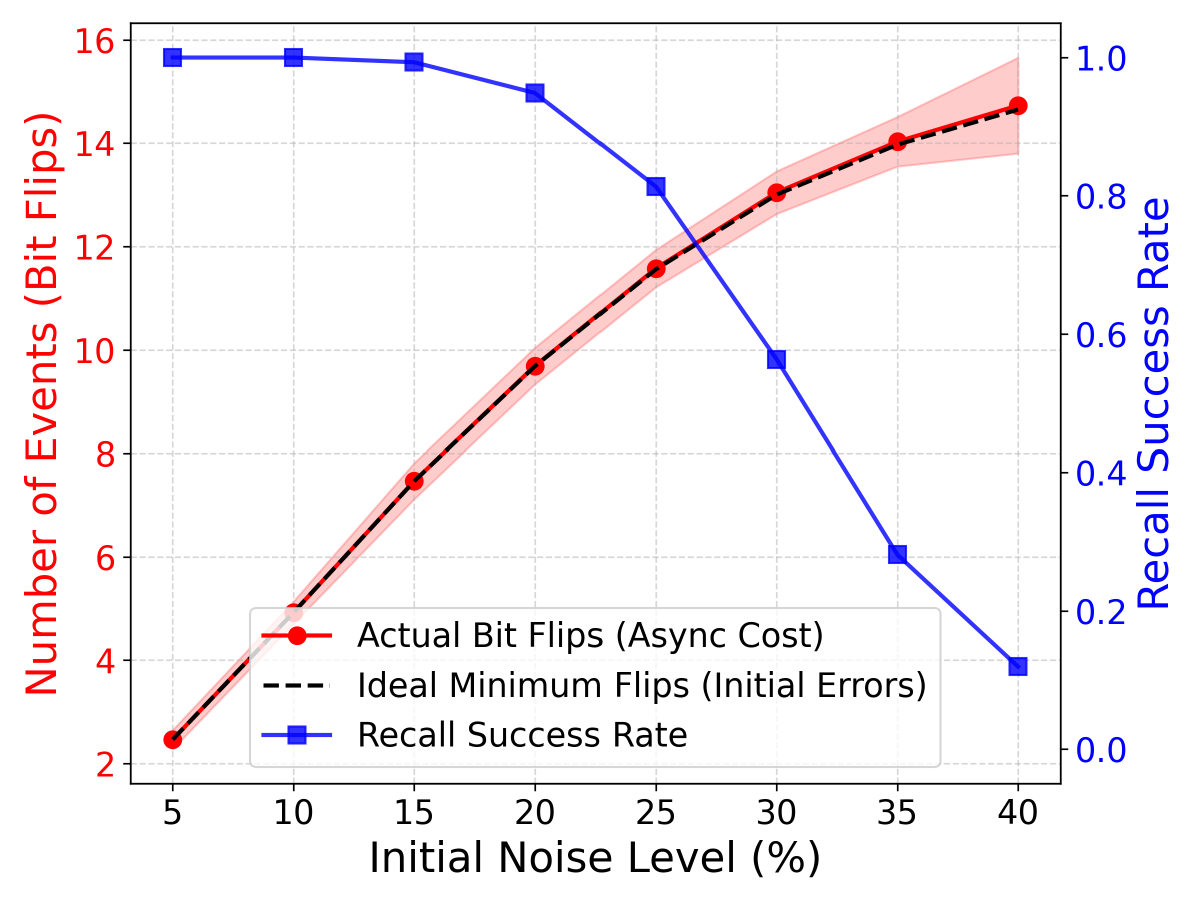}
\caption{Event-driven retrieval efficiency under asynchronous
  updates. The total number of state transitions (bit flips) required
  for convergence is plotted against the initial noise level (red
  solid line), compared with the theoretical minimum number of flips
  corresponding to the number of initial errors (black dashed
  line). The secondary y-axis shows the recall success rate (blue
  squares). Shaded areas indicate standard deviation over 50 trials.}
\label{fig:event_efficiency}
\end{figure}

Figure~\ref{fig:event_efficiency} plots the average number of actual
bit flips (red solid line) and the ideal minimum flips (black dashed
line) as a function of the initial noise level, for a network of
$N=50$ neurons storing $P=150$ patterns ($\gamma=0.1$).
The results indicate a close alignment between the actual and ideal
event counts across all tested noise levels (from 5\% to 40\%). Within
the robust regime governed by $\gamma=0.1$, the network appears to
correct erroneous bits directly, without inducing many secondary bit
flips. For instance, even at a high noise level of 40\%, where the
initial state contains approximately 15 erroneous bits on average, the
asynchronous dynamics converge in an equally minimal number of events
(around 15 flips), albeit with a lower recall success rate ($< 10\%$,
blue line) due to the reduced basin size relative to the noise
magnitude. At a lower, more practical noise level of 20\%
(approximately 10 initial errors), the network achieves a high recall
success rate of roughly 95\% using exactly the theoretical minimum of
10 events.

This efficient, minimal-event trajectory suggests that, under the
tested conditions (random patterns and $\gamma=0.1$), the energy
landscape of the KLR network is characterized by deep, smooth
attractor basins largely free of pronounced rugged local minima. From
an engineering perspective, this suggests that an event-driven
implementation of the KLR network can require significantly fewer
computations per retrieval compared to a synchronous implementation,
which must evaluate all $N$ neurons at every time step. For example,
retrieving a pattern from 20\% initial noise in a 50-neuron network
typically takes about 3 steps using synchronous updates (as seen in
Fig.~\ref{fig:retrieval_dynamics}), requiring $3\times 50=150$ state
evaluations. In contrast, the event-driven asynchronous update
achieves the same retrieval using only about 10 evaluations
(corresponding to the 10 initial bit errors).  We observed that this
near-optimal efficiency is not limited to small networks but scales to
larger systems. As detailed in
Appendix~\ref{app:scalability_efficiency}, identical experiments
conducted on networks of size $N=100$ and $N=200$ showed a similar
alignment between actual and ideal event counts.

\section{Discussion}
\label{sec:discussion}
The empirical results presented in this study demonstrate that KLR
Hopfield networks achieve high storage capacity while also exhibiting
retrieval dynamics that are well suited for asynchronous, event-driven
computation. In this section, we discuss the geometric and practical
implications of these findings.

\subsection{Margin-Induced Smoothness of the Attractor Landscape}
The empirical similarity between synchronous and asynchronous
retrieval trajectories (Sec.~\ref{sec:similarity}), together with the
high event efficiency (Sec.~\ref{sec:efficiency}), provides insight
into the energy landscape of KLR networks. In classical Hopfield
networks, asynchronous updates are often required to guarantee
convergence to a local minimum and to avoid oscillatory behavior in
synchronous updates~\cite{Hopfield1982}. However, even with
asynchronous updates, classical networks often traverse rugged
landscapes, potentially falling into spurious states or requiring many
state transitions before reaching a memory basin.

In contrast, the energy landscape sculpted by KLR learning, under
appropriate kernel parameters, appears to be characterized by smooth
and deep basins of attraction. The observation that asynchronous
dynamics correct erroneous bits with a minimal number of secondary bit
flips (Fig.~\ref{fig:event_efficiency}) suggests a lack of rugged
local minima or saddle points along the retrieval paths within these
basins.

Mechanistically, this robust alignment between synchronous and
asynchronous trajectories can be attributed to the large
classification margins induced by the KLR optimization process. In
classical Hopfield networks trained with Hebbian learning, the local
field magnitudes (margins) are typically small and exhibit high
variance, making the update direction sensitive to the specific
sequence of prior bit flips. Conversely, by aggressively maximizing
the margin in the high-dimensional feature space, KLR ensures that the
local field $h_i (\bm{s})$ for each stored pattern is strong and
consistently aligned with the target state. We hypothesize that this
large margin effectively minimizes the state perturbations caused by
asynchronous update order permutations. Consequently, the gradient of
the landscape dictates the trajectory, rendering the difference
between batch and sequential evaluations small in practice. This
margin-induced smoothness provides a plausible explanation for why the
network can simultaneously support high capacity ($P/N \approx 30$)
and efficient, direct retrieval trajectories.

\subsection{The Trade-off Between Capacity and Locality}
Our experiments highlight the critical role of the kernel locality
parameter $\gamma$. In our previous work, we identified a highly
localized regime ($\gamma = 0.02$) that maximizes the ``sharpness''
(depth) of individual attractors, and is optimal for static memory
retrieval under minimal noise~\cite{tamamori_letter, tamamori_nolta_a}.
However, the present study indicates that achieving high storage loads
($P/N > 20$) while maintaining robustness against significant initial
noise (e.g., 10--20\%) requires a broader kernel ($\gamma = 0.1$).

This observation suggests a fundamental geometric trade-off: a smaller
$\gamma$ isolates patterns and suppresses interference but reduces the
effective basin size, whereas a larger $\gamma$ enlarges the basins
and improves noise robustness but increases the risk of crosstalk at
high storage loads. The ability to tune $\gamma$ allows the KLR
network to adapt to different task requirements, providing flexibility
beyond that of standard dot-product associative memories.

\subsection{Implications for Neuromorphic Hardware}
The demonstrated event-driven efficiency suggests that KLR
Hopfield networks are well suited for emerging neuromorphic
hardware architectures, such as SNNs implemented on chips like
Loihi~\cite{Davies2018} or TrueNorth~\cite{Merolla2014}.

In digital hardware implementations of Hopfield networks, a trade-off
typically exists between execution speed and dynamical
stability. Synchronous updates evaluate all $N$ neurons in a single
step, offering high throughput, but they lack formal convergence
guarantees and are susceptible to
oscillations~\cite{Bruck1990}. Conversely, asynchronous updates
guarantee convergence to stable states but require $N$ sequential
evaluations per epoch, generally leading to higher latency.

The event-driven approach analyzed in this study offers a potential
mitigation for this trade-off. In a traditional synchronous
implementation, evaluating the dense kernel product for all $P$
patterns and $N$ neurons results in a computational complexity of
$\mathcal{O}(PN)$ per step. By leveraging the asynchronous,
event-driven scheme, calculations are triggered only when a neuron's
state actually flips. As shown in Fig.~\ref{fig:event_efficiency},
correcting an initial 20\% error in a 50-neuron network requires
approximately 10 discrete event evaluations. Under such optimal
retrieval conditions, the required number of updates scales with the
initial error count rather than the network size $N$. This reduction
in redundant state evaluations can alleviate the latency penalty
typically associated with asynchronous systems, supporting its
viability for scalable, low-power neuromorphic architectures.

\subsection{Limitations and Future Work}
While this study empirically demonstrates the efficiency of
asynchronous KLR networks, several limitations remain. First, our
analysis is based on uncorrelated random binary patterns with
relatively small network sizes ($N\leq 200$).  This choice was
deliberate, as this scale facilitates a rigorous, large-ensemble
analysis while random patterns provide a standardized,
maximum-interference benchmark for evaluating empirical capacity
limits ($P/N \approx 30$) and probing the boundaries of dynamical
stability. However, real-world data (e.g., images or language
embeddings) typically involve much larger network dimensions ($N$) and
are commonly assumed to lie on or near low-dimensional manifolds
embedded in high-dimensional ambient spaces, often exhibiting strong
structural correlations~\cite{Fefferman2016, Ghojogh2023}. Our
previous study showed that kernel methods can exploit such
correlations to achieve higher practical storage
loads~\cite{tamamori_2026a}. Verifying whether the minimal-event
asynchronous trajectories observed here persist when scaled up to
standard benchmarks like MNIST or CIFAR datasets, which introduce both
high dimensionality and skewed energy landscapes induced by
structured, correlated datasets remains an important direction for
future work.

Second, while the event-driven scheme substantially reduces the number
of state updates, the computational cost of evaluating the RBF kernel
function itself,
$K(\bm{s}, \boldsymbol{\xi}^\mu) = \exp(-\gamma \|\bm{s} -
\boldsymbol{\xi}^\mu\|^2)$, remains non-negligible. Computing exact
Euclidean distances and exponential functions poses challenges for
resource-constrained, spike-based neuromorphic hardware. To fully
realize the hardware efficiency discussed in this study, future work
should explore hardware-friendly kernel approximations. Potential
approaches include polynomial kernels (which require only
multiply-accumulate operations)~\cite{Schölkopf2001}, piecewise linear
approximations of the exponential function, and extreme low-precision
weight quantization, which we previously showed to be effective in
maintaining KLR capacity~\cite{tamamori_2026b}. Combining these
approaches with the proposed event-driven framework is an important
step toward practical deployment on edge devices.

Finally, a rigorous mathematical proof guaranteeing global Lyapunov
stability for the asymmetric effective weight matrices in KLR remains
an open theoretical problem, although a margin-based mechanistic
explanation is provided in Appendix~\ref{app:energy_landscape}.
Moreover, while our empirical results demonstrate storage loads of
$P/N \approx 30$, a large-scale quantitative comparison with modern
continuous-state architectures, such as exponential Hopfield
networks~\cite{Demircigil2017} and Transformer-based associative
memories, remains an important direction for future work.
Establishing the theoretical trade-offs between the margin-induced
capacity of KLR and the energy-function-induced capacity of MHNs will
help clarify design principles for next-generation associative memory
systems.

\section{Conclusion}
\label{sec:conclusion}
In this study, we investigated the feasibility and efficiency of
asynchronous retrieval dynamics in high-capacity Kernel Logistic
Regression (KLR) Hopfield networks. Our empirical analysis
demonstrated that, under the tested conditions, asynchronous updates
do not degrade the performance of KLR networks; instead, they provide
a functionally similar and efficient alternative to synchronous,
batch-level processing.

Specifically, our experiments yielded three main findings. First, the
convergence trajectories of synchronous and asynchronous updates are
closely aligned, suggesting that the KLR energy landscape is robust to
sequential perturbations. Second, by tuning the kernel locality
parameter ($\gamma = 0.1$), the asynchronous network achieves an
empirical storage capacity approaching $P/N \approx 30$ for random
patterns while maintaining high recall accuracy under noisy initial
conditions. Third, we quantified the computational cost of the
asynchronous scheme. The network corrects noisy initial states using a
number of bit flips that closely matches the initial Hamming distance,
indicating a direct trajectory to the target attractor basin.

These results suggest that the large-margin attractors formed by KLR
learning suppress spurious oscillations commonly associated with
recurrent dynamics. By transforming dense, synchronous matrix
operations into sparse, event-driven state transitions, KLR Hopfield
networks can achieve substantial reductions in computational and
energy costs. This work suggests that KLR-based architectures provide
a promising approach for implementation on next-generation,
event-driven neuromorphic hardware, bridging the gap between
high-capacity associative memory models and practical low-power
artificial intelligence systems.

\funding
Not applicable.

\conflictsofinterest
The author declares no competing interests.

\authorcontribution
The sole author contributed to the present work.

\aitools
The author used ChatGPT (GPT-5.3) and Gemini 2.5 Pro for proofreading the English
manuscript.

\appendix

\section{Mechanistic Explanation of Asynchronous Stability}
\label{app:energy_landscape}

In classical Hopfield networks with symmetric weights
($w_{ij} = w_{ji}$), the asynchronous update rule strictly minimizes
the Ising energy $E(\bm{s}) = -\frac{1}{2} \sum_{i,j} w_{ij} s_i
s_j$. In contrast, our KLR-trained network utilizes a pseudo-energy
function $V(\bm{s})$, defined in Eq.~(\ref{eq:pseudo_energy}), where
the effective weight matrix derived from the kernel expansion is
generally asymmetric. Consequently, $V(\bm{s})$ is not guaranteed to
be a strict Lyapunov function globally.

Nevertheless, our empirical results demonstrate convergence without
oscillations under asynchronous updates. This behavior can be
explained semi-formally by considering the large classification
margins induced by KLR optimization.

Consider an asynchronous update at step $t$, where a single neuron $i$
is selected. Suppose its state flips from $s_i$ to
$s_i^{\text{new}} = -s_i$. By the update rule
$s_i^{\text{new}} = \mathrm{sign}(h_i(\bm{s}))$, this flip occurs only
when $s_i h_i(\bm{s}) < 0$. Let us analyze the change in the
pseudo-energy,
$\Delta V = V(\bm{s}^{\text{new}}) - V(\bm{s})$. This change can be
decomposed into two components:
\begin{equation}
    \Delta V = \Delta V_{\text{local}} + \Delta V_{\text{cross}},
\end{equation}
where $\Delta V_{\text{local}}$ represents the contribution from
neuron $i$ aligning with its local field, and
$\Delta V_{\text{cross}}$ represents the interference caused by the
change in $s_i$ on the local fields $h_j$ of all other neurons
$j \neq i$.

The local energy change is strictly negative, and its magnitude is
proportional to the local field (margin):
\begin{equation}
  \Delta V_{\text{local}} = - (s_i^{\text{new}} - s_i) h_i(\bm{s}) = - 2 |h_i(\bm{s})| < 0.
\end{equation}
The cross-talk interference term arises because flipping $s_i$
perturbs the kernel evaluations $K(\bm{s}, \boldsymbol{\xi}^\mu)$
for all stored patterns. The total interference is bounded as:
\begin{equation}
    |\Delta V_{\text{cross}}| = \left| \sum_{j \neq i} s_j \left( h_j(\bm{s}^{\text{new}}) - h_j(\bm{s}) \right) \right| \le I_{\max},
\end{equation}
where $I_{\max}$ denotes the maximum cross-neuron interference
induced by a single bit flip. To derive a heuristic upper bound for
$I_{\max}$, we exploit properties of the RBF kernel. Since the RBF
kernel is bounded ($0 < K(\cdot, \cdot) \le 1$), the change in any
local field $\Delta h_j = h_j(\bm{s}^{\text{new}}) - h_j(\bm{s})$
induced by a single bit flip is finite and bounded by the learned
dual variables:
\begin{align}
  |\Delta h_j| &\le \sum_{\mu=1}^P |\alpha_{\mu j}|
  \left| K(\bm{s}^{\text{new}}, \boldsymbol{\xi}^\mu) -
  K(\bm{s}, \boldsymbol{\xi}^\mu) \right| \nonumber\\
  &\le \sum_{\mu=1}^P |\alpha_{\mu j}|.
\end{align}
Summing over all $j \neq i$ yields a conservative upper bound,
$I_{\max} \le \sum_{j \neq i} \sum_{\mu=1}^{P} |\alpha_{\mu j}|$.

For the asynchronous dynamics to monotonically decrease the
pseudo-energy ($\Delta V < 0$) and ensure convergence without
oscillations, the following sufficient condition must hold:
\begin{equation}
  2 |h_i(\bm{s})| > I_{\max}.
  \label{eq:stability_condition}
\end{equation}
In standard associative memories, the margin $|h_i|$ is often small,
and a single flip can violate this condition, leading to energy
increases and limit cycles. In contrast, KLR explicitly maximizes the
margin $|h_i|$ in the high-dimensional feature space. In the robust
operating regimes (e.g., $\gamma = 0.1$), KLR drives the local field
magnitudes to be sufficiently large such that the bounded cross-talk
interference $I_{\max}$ becomes negligible. As a result, condition
(\ref{eq:stability_condition}) is satisfied over most of the attractor
basins. This margin dominance ensures that local, energy-decreasing
flips govern the global dynamics, leading to the efficient,
oscillation-free trajectories observed in our experiments.

\section{Scalability of Retrieval Dynamics}
\label{app:scalability_dynamics}

In Section~\ref{sec:similarity}, we demonstrated the empirical
similarity between synchronous and asynchronous retrieval dynamics
using a network of $N = 50$ neurons
(Fig.~\ref{fig:retrieval_dynamics}). To verify that this equivalence
extends to larger systems and is not a finite-size artifact, we
conducted the same experiments using networks of size $N = 100$ and
$N = 200$.

\begin{figure}[t]
    \centering
    \includegraphics[width=0.49\textwidth]{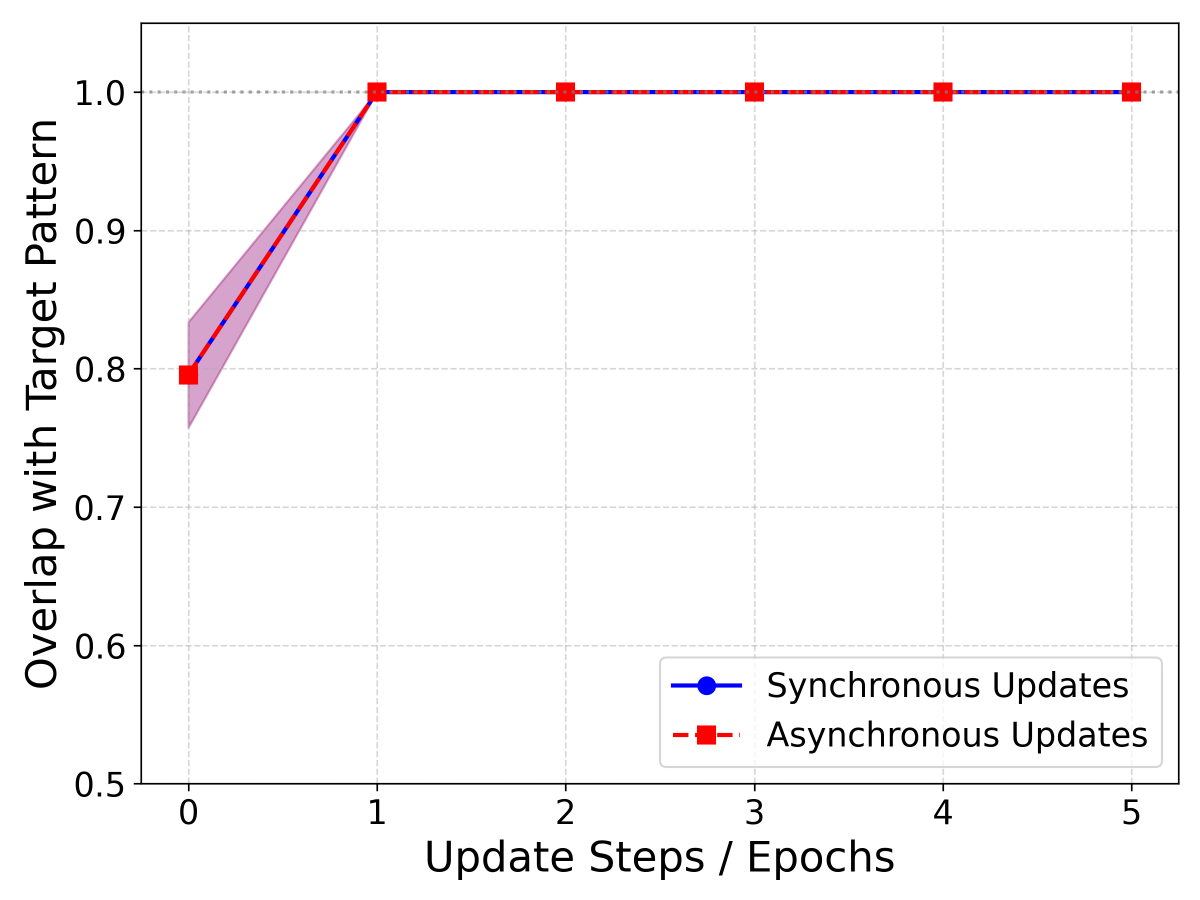}
    \vspace{3mm}
    (a) $N=100$
    \includegraphics[width=0.49\textwidth]{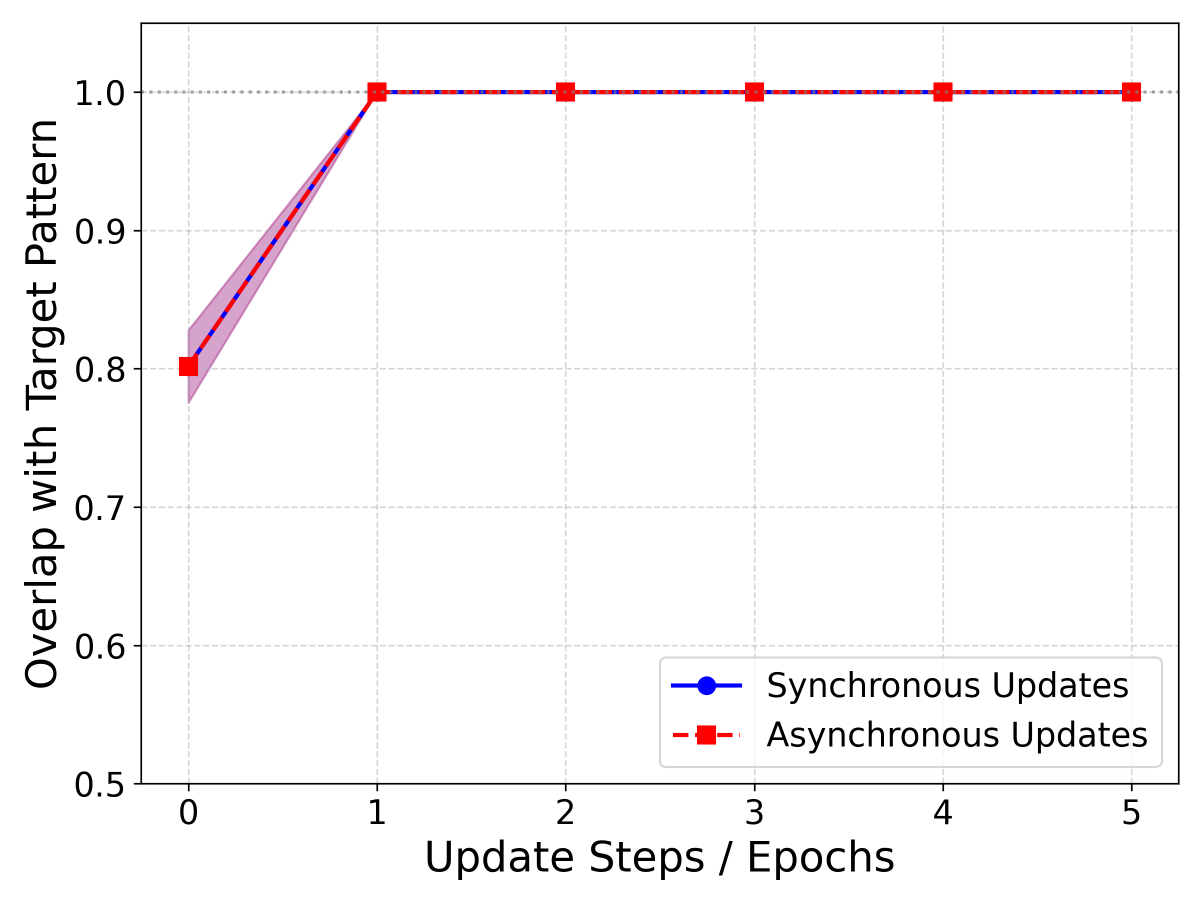}
    (b) $N=200$
    \caption{Retrieval dynamics at larger network sizes. The
      plots compare the convergence trajectories of synchronous and
      asynchronous updates for networks of size (a) $N=100$ and (b)
      $N=200$. The experimental conditions ($P/N=3.0$, $\gamma=0.1$,
      20\% initial noise) and plotting conventions are identical to
      those in Fig.~\ref{fig:retrieval_dynamics}. In both cases, the
      synchronous and asynchronous trajectories are empirically
      indistinguishable, confirming the scalability of the observed
      behavior.}
    \label{fig:retrieval_dynamics_largeN}
\end{figure}

As in the main text, the networks were trained at a storage load of
$P/N = 3.0$ with $\gamma = 0.1$. The retrieval process was initiated
from states with 20\% random noise.
Figure~\ref{fig:retrieval_dynamics_largeN} shows the average overlap
trajectories over 50 independent trials.

The results show that the close alignment between synchronous (blue
solid lines) and asynchronous (red dashed lines) updates persists as
the network size increases. The trajectories approach perfect recall
(overlap $\approx 1.0$) more closely for larger $N$, which can be
attributed to the increased orthogonality of random patterns in
higher-dimensional spaces. The overlapping error bands indicate that
the choice of update scheme does not introduce significant deviations
in the retrieval path, further supporting the robustness of the KLR
energy landscape across different scales.

\section{Scalability of Event-Driven Efficiency}
\label{app:scalability_efficiency}

\begin{figure}[t]
    \centering
    \includegraphics[width=0.49\textwidth]{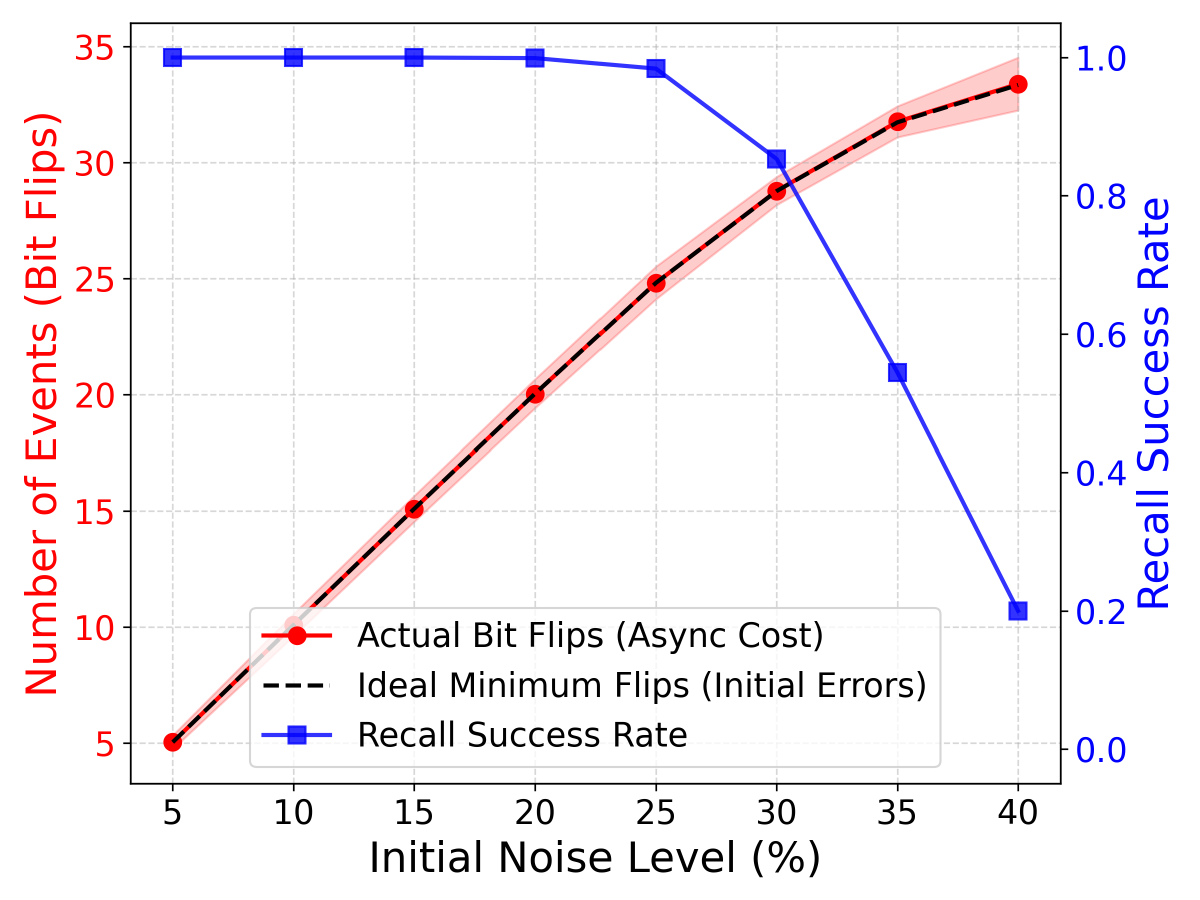}
    \vspace{3mm}
    (a) $N=100$
    \includegraphics[width=0.49\textwidth]{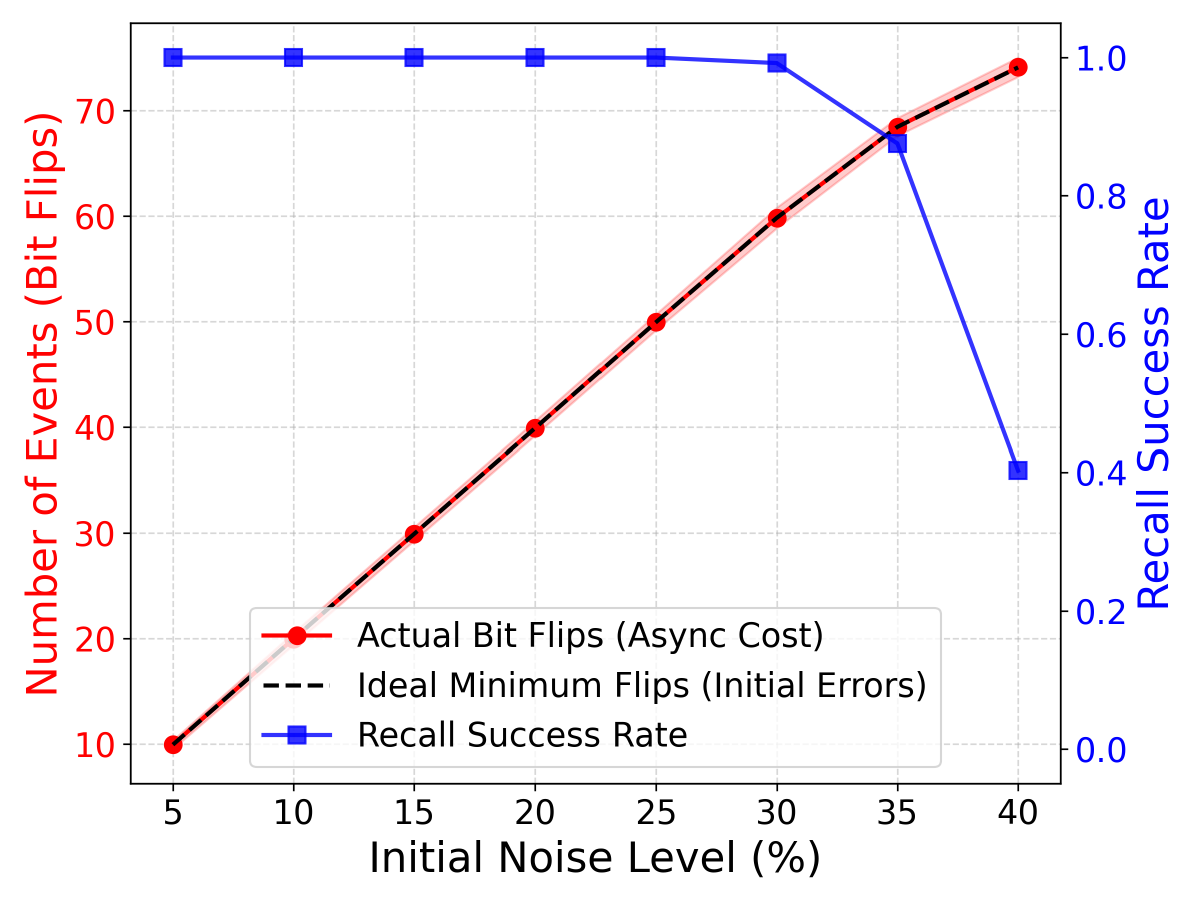}
    (b) $N=200$
    \caption{Event-driven efficiency at larger network sizes.
      The plots compare the actual number of bit flips (red) to the
      ideal minimum (black dashed) for networks of size (a) $N=100$
      and (b) $N=200$. The experimental conditions
      ($P/N=3.0, \gamma=0.1$) and plotting conventions are identical
      to those in Fig.~\ref{fig:event_efficiency}. In both cases, the
      actual event count closely follows the theoretical minimum
      across all tested noise levels, confirming the scalability of
      the efficient retrieval dynamics.}
    \label{fig:event_efficiency_largeN}
\end{figure}

In Section~\ref{sec:efficiency}, we showed that the asynchronous
KLR network follows an efficient, minimal-event trajectory during
retrieval using a network of $N = 50$ neurons
(Fig.~\ref{fig:event_efficiency}). To examine potential finite-size
effects and to verify the scalability of this behavior, we repeated
the event-counting experiment for larger network sizes: $N = 100$
and $N = 200$.

The networks were trained under the same operating regime
($\gamma = 0.1$) with a fixed storage load of $P/N = 3.0$. We measured
the total number of bit flips required for convergence across various
initial noise levels and compared it with the ideal minimum (the
initial Hamming distance).

Figure~\ref{fig:event_efficiency_largeN} presents the results. The
alignment between the actual bit flips (red solid lines) and the
ideal minimum (black dashed lines) is preserved for both $N = 100$
and $N = 200$. The network corrects erroneous bits directly, without
inducing secondary oscillations or unnecessary state transitions,
across all tested conditions. Furthermore, the recall success rate
(blue squares) remains high, maintaining near 100\% accuracy at
20\% noise for both network sizes.

These results suggest that the smooth, large-margin geometry of the
KLR attractor basins extends to larger system sizes. The scalability
of this minimal-event behavior further supports the suitability of
KLR-trained networks for large-scale implementation on event-driven
neuromorphic hardware.

\end{document}